\documentclass[runningheads]{llncs}

\usepackage{graphicx}
\usepackage{multirow}
\usepackage{booktabs}
\usepackage{amsmath}
\usepackage{amssymb}

\usepackage{bbding}
\usepackage{hyperref}
\newcommand{\etal}{\textit{et al.}}
\begin{document}
\title{Deep Semi-supervised Knowledge Distillation for Overlapping Cervical Cell Instance Segmentation}
%
%
\author{Yanning Zhou\inst{1},
Hao Chen\inst{1},
Huangjing Lin\inst{1},
Pheng-Ann Heng\inst{1,2}}
%
\authorrunning{Y. Zhou et al.}

\institute{$^1$Department of Computer Science and Engineering, The Chinese University of Hong Kong, Hong Kong SAR, China\\
\email{\{ynzhou,hchen,hjlin,pheng\}@cse.cuhk.edu.hk}\\
$^2$Guangdong Provincial Key Laboratory of Computer Vision and Virtual Reality Technology, Shenzhen Institutes of Advanced Technology, Chinese Academy of Sciences, Shenzhen, China}

\maketitle              
\begin{abstract}
Deep learning methods show promising results for overlapping cervical cell instance segmentation. 
However, in order to train a model with good generalization ability, voluminous pixel-level annotations are demanded which is quite expensive and time-consuming for acquisition. 
In this paper, we propose to leverage both labeled and unlabeled data for instance segmentation with improved accuracy by knowledge distillation. 
We propose a novel Mask-guided Mean Teacher framework with Perturbation-sensitive Sample Mining (MMT-PSM), which consists of a teacher and a student network during training.
Two networks are encouraged to be consistent both in feature and semantic level under small perturbations. 
The teacher's self-ensemble predictions from $K$-time augmented samples are used to construct the reliable pseudo-labels for optimizing the student.
We design a novel strategy to estimate the sensitivity to perturbations for each proposal and select informative samples from massive cases to facilitate fast and effective semantic distillation.
In addition, to eliminate the unavoidable noise from the background region, we propose to use the predicted segmentation mask as guidance to enforce the feature distillation in the foreground region.
Experiments show that the proposed method improves the performance significantly compared with the supervised method learned from labeled data only, and outperforms state-of-the-art semi-supervised methods. 
Code: \url{https://github.com/SIAAAAAA/MMT-PSM}

\end{abstract}

\section{Introduction}
Pap smear test is the recommended procedure for earlier cervical cancer screening worldwide~\cite{Pap1942}. 
By estimating the cell type and the cytological features, e.g., nuclei size, nuclear cytoplasmic ratio and multi-nucleation, it provides clear guidance for clinical management and further treatment~\cite{solomon20022001}.
Automatic cervical cell segmentation can free doctors from time-consuming work and reduce the intra-/inter-observer variability~\cite{gencctav2012unsupervised,lu2015improved,song2017accurate,zhou2019irnet}. 
Specifically, Deep Learning (DL) methods show promising results for cell nuclei segmentation~\cite{al2018deep,RAZA2019160,zhou2019irnet}.
However, optimizing the DL methods heavily relies on numerous data with expensively dense annotations by experts, which limits the model to acquire higher accuracy and better generalization ability.
Since unlabeled data is easily accessible, how to leverage both limited labeled and large amounts of unlabeled data raises researchers' attention to improve the performance further for medical image analysis~\cite{wu2020}.

Several works have been done in medical image community for Semi-Supervised Learning (SSL) on classification and segmentation~\cite{su2015interactive,meier2014patient,bai2017semi,baur2017semi,nie2018asdnet,yu2019uncertainty,cui2019semi,su2019local}. 
Bai \etal~\cite{bai2017semi} proposed a self-training strategy by alternatively assigning labels to unlabeled data and optimizing the model parameters.
Nie \etal~\cite{nie2018asdnet} introduced an adversarial learning training strategy by selecting informative regions in unlabeled data to train the segmentation network.
Shi \etal~\cite{shi2020graph} created more reliable ensemble targets for feature and label predictions via the graph to encourage features mapped in the same cluster being more compact.
Knowledge distillation~\cite{hinton2015distilling}, which was first used in model compression by encouraging the small model to mimic the behavior of a deeper model, has demonstrated excellent improvements mostly for classification setups~\cite{romero2014fitnets,che2015distilling,tarvainen2017mean} and shown the potential benefit for semi-supervised learning~\cite{tarvainen2017mean} and domain adaptation~\cite{ge2020mutual}.
Chen \etal~\cite{chen2017learning} extended it to the detection scenario with proposal-based method, and presented to learn a compact detector by distilling from both features and predictions. 
However, directly using entire feature maps will inevitably introduce the noise from the background.
To eliminate the noise in background, Wang \etal~\cite{wang2019distilling} conducted feature distillation within the region close to objects based on prior knowledge.
Other approaches~\cite{jeong2019consistency,cai2019exploring} added consistent regularization either in region-based or relation-based.
Although achieving promising progress, they do not consider the informative degree for each sample, which is one of the bottlenecks for further improving the performance. 
In medical imaging, researchers attempted to apply knowledge distillation to segmentation problems.
Wang \etal~\cite{wang2019segmenting} employed the teacher student network in 3D optical microscope images via knowledge distillation.
Another approach ~\cite{yu2019uncertainty} introduced uncertainty estimation into knowledge distillation for 3D left atrium segmentation.
Instance segmentation, however, is a more challenging task that requires an additional detection step to distinguish the individual instances~\cite{he2017mask}.
The potential of the knowledge distillation has not been well explored on it.

In this paper, we propose a novel deep semi-supervised knowledge distillation framework 
called Mask-guided Mean Teacher with Perturbation-sensitive Sample Mining (MMT-PSM) for overlapping cervical cell instance segmentation, which conducts both semantic and feature distillation.
The proposed end-to-end trainable framework consists of a teacher model and a student model under the same backbone.
Given a sample with different small perturbations, the proposed method encourages the predictions from two networks being consistent.
The mean prediction of the $K$-time augmented samples from the teacher network are considered as the pseudo-label to supervise the student network. 
A perturbation-sensitive sample mining strategy is used to resolve the meaningless guidance from easy cases in unbalanced and massive data.
Furthermore, we propose the mask-guided feature distillation which encourages the feature consistency only for the foreground region to alleviate the side effect in the noisy background.
We perform comprehensive evaluation on cervical cell segmentation task.
Results indicate that the proposed algorithm significantly improves the instance segmentation accuracy, consistently across different numbers of labeled data, and also outperforms other state-of-the-art semi-supervised methods.

\section{Method}

\begin{figure}[!h]
\centering

	\includegraphics[width=0.95\textwidth]{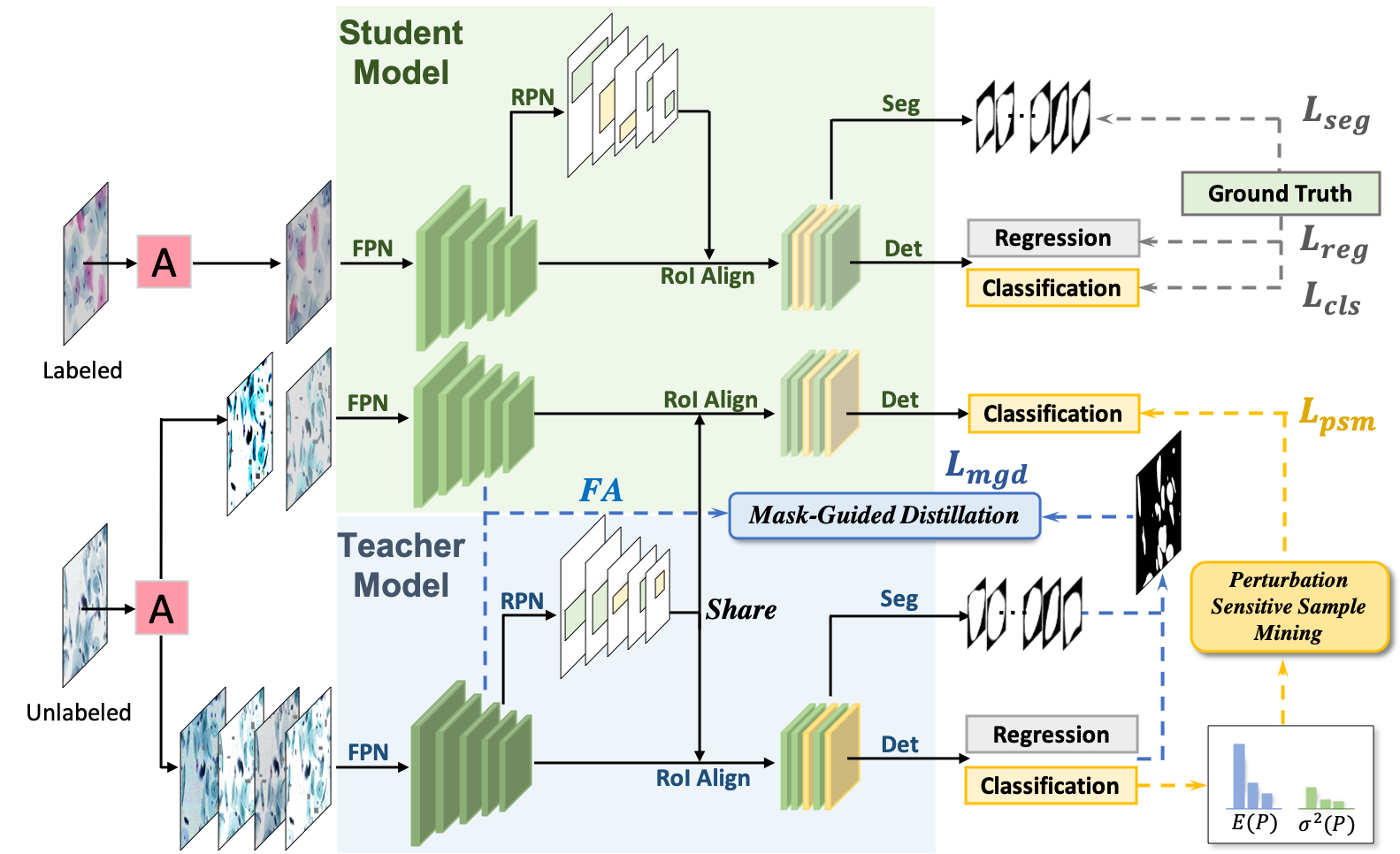}

	\caption{ Overview of the proposed framework.
	Annotated data is passed through the student to calculate standard supervised losses. 
    Meanwhile, $K$-time data augmentation (A) is applied to the unlabeled data, which are fed through the teacher to generate soft pseudo-labels to optimize the student. 
	The perturbation sensitivity is estimated on the $K$ teacher's predictions to select samples for optimizing the student.
	The predicted masks are used as a guidance for feature distillation (i.e., Mask-Guided Distillation). FA denotes the Feature Adaptation layer.}
		\label{framework}
	
\end{figure}

\subsection{Mean Teacher Framework for Instance Segmentation.}

Formally, let $\mathcal{D}_{L} = \left \{ \left ( x_{i},y_{i} \right ) \right \}_{i=1}^{N}$ denote the labeled set and $\mathcal{D}_{U} = \left \{ \left ( x_{i} \right ) \right \}_{i=N+1}^{N+M}$ denote the unlabeled set. 
The goal of semi-supervised learning is to improve the performance by leveraging the hidden information in $\mathcal{D}_{U}$. 
In this work, we adopt Mask R-CNN~\cite{he2017mask} as the instance segmentation model for both the student and the teacher, which consists of four modules: 
1) A shared Feature Pyramid Network (FPN) extracts features as inputs for the other modules, 2) a Region Proposal Network (RPN) equipped with RoI Align layer to generate the object proposals, 3) a detection branch (Det) and 4) a segmentation branch (Seg) which take features and proposals as inputs, and then predict the detection scores, the spacial revision vectors and the segmentation results, respectively. 
We use Mean Teacher algorithm (MT)~\cite{tarvainen2017mean} as our basic framework, which consists of a teacher and a student model sharing the same architecture and encourages the predictions being consistent under small perturbations. 
Instead of optimizing the teacher by SGD, exponential moving average (EMA) weight in the student is used to form a better teacher model~\cite{tarvainen2017mean}: ${\theta}'_{t} = \alpha{\theta}'_{t-1} +(1-\alpha) {\theta}_{t-1}$, where ${\theta}'_{t-1}$ and ${\theta}_{t-1}$ are the teacher’s and student's weights in $(t-1)$ step, and $\alpha$ controls the updating speed.

\subsection{Perturbation-sensitive Samples Distillation.}
One difficulty in applying MT on the instance segmentation is the sample-imbalanced problem in proposals.
Directly computing loss on all predictions is not effective because most samples lie in background regions and can be easily distinguished, which overwhelms the useful information. 
We propose to use the mean predictions of $K$-time augmented samples as more reliable targets from the teacher and select samples based on its sensitivity to perturbations.

\noindent\textbf{Self-ensembling pseudo-label.}
Specifically, for each image $x_{i}$ in $\mathcal{D}_{U}$, a stochastic Augmentor (A) is used to augment $K$ samples $\left \{ x_{i,1}^{\mathcal{T}},\dots,x_{i,K}^{\mathcal{T}} \right \}$ for the teacher and $L$ augmented samples $\left \{ x_{i,1}^{\mathcal{S}},\dots, x_{i,L}^{\mathcal{S}}\right\}$ for the student.
To acquire the same candidates for further loss calculation between networks, the proposals $ \mathcal{R}_{x_{i}^{\mathcal{T}}}$ generated from teacher's RPN are shared for both teacher and student.
Self-ensemble predictions from a collection of augmented data have been considered as a more reliable target in classification~\cite{berthelot2019mixmatch}.
Here, we calculate the average predictions across $K$ augmented samples to generate the soft pseudo-label in teacher network:

\begin{equation}
  \overline{P}_{i} = \frac{1}{K}\sum_{k=1}^{K}f_{cls}\left (x_{i,k}^{\mathcal{T}},\mathcal{R}_{x_{i}^{\mathcal{T}}} ; {\theta}'  \right ).
\end{equation}
$f_{cls}\left (\cdot ; {\theta}'  \right )$ denotes the classification sub-branch in teacher Det.
A sharpen function $S(\overline{P}_{i})=\overline{P}_{i}^{t}/\sum_{j=1}^{c}\overline{P}_{j}^{t}$ is further used to implicitly achieve entropy minimization~\cite{berthelot2019mixmatch}, in which $c$ denotes the number of categories. We set $t=0.5$ in our study.
See the supplementary material for augmentation details and ablation study of the  temperature. 

\noindent\textbf{Perturbation-sensitive sample mining.}
We hypothesize that perturbation-sensitive samples, which have larger prediction accuracy gaps between teacher and student, are more informative and beneficial for training. 
Firstly, the class with the maximum categorical probability in the self-ensembling prediction is assigned as its hard pseudo-label.
Then we calculate the variance among $K$ augmented samples as its degree of perturbation sensitivity:
\begin{equation}
   Var(x_{i}^{\mathcal{T}}) = \frac{1}{K}\sum_{k=1}^{K} \left ( f_{cls}\left (\hat{x}_{i,k}^{\mathcal{T}},\mathcal{R}_{x_{i}^{\mathcal{T}}} ; {\theta}'  \right )- \overline{P}_{i} \right ) ^{2}.
\end{equation}
All samples whose hard pseudo-labels are foreground classes remain.
Meanwhile, background samples are sorted by descending according to the variances and kept the Top-$s$, where $s$ is the number of foreground samples. 
The perturbation-sensitive sample mining loss $\mathcal{L}_{psm}$ is calculated on the selected samples as follow:
\begin{equation}
   \mathcal{L}_{psm} =\frac{1}{ML}\sum _{i=N+1}^{N+M}\sum _{l=i}^{L}w\mathcal{L}_{ce}\left(f_{cls}\left(x_{i,l}^{\mathcal{S}} ,\mathcal{R}_{x_{i}^{\mathcal{T}}} ; {\theta}\right),S(\overline{P}_{i}) \right).
\end{equation}
$f_{cls}\left (\cdot ; \theta  \right )$ denotes the classification sub-branch in student Det, $\mathcal{R}_{x_{i}^{\mathcal{T}}}$ and $\overline{P}_{i}$ denote the proposals and soft pseudo-labels for remained perturbation-sensitive samples, and $\mathcal{L}_{ce}$ is the cross-entropy loss.
$w$ is a class-balanced weight and is set empirically as $1.5$ for the background and 1 for others.

\subsection{Mask-Guided Feature Distillation.}

Study shows~\cite{romero2014fitnets} that intermediate representations from the teacher can also improve the training process and final performance of the student in the classification task.
However, directly minimize the difference in entire feature maps could harm the performance since it would introduce the noise in the background region. 
Therefore, we design to force the student only mimicking the teacher under the guidance of semantic segmentation results. 

Firstly, an adaptation layer is added after each output stage of FPN, which is proved to be advantageous for feature distillation~\cite{romero2014fitnets}.
Here we use a $1\times1$ convolution as the adaptation layer and reduce the input feature dimension by half. 
Then the instance masks and bounding box's locations from the teacher are used to generate binary semantic masks.
Let $Z_{tijc}^{\mathcal{S}}$ and $Z_{tijc}^{\mathcal{T}}$ denotes the student's and the teacher's feature value in the $c$-th channel at location $(i,j)$ from the adaptation layer after FPN's $t$-th stage. 
We aim to encourage the consistency by minimizing feature distance through the mask-guided distillation loss:
\begin{equation}
\mathcal{L}_{mgd} = \frac{1}{CT}\sum^{T}_{t=1}\frac{\sum^{W}_{i=1}\sum^{H}_{j=1}\sum^{C}_{c=1}M_{tij}\circ\left \|  Z_{tijc}^{\mathcal{T}} -Z_{tijc}^{\mathcal{S}} \right \|_{2}^{2}}{\sum^{W}_{i=1}\sum^{H}_{j=1}M_{tij}} .
\end{equation}
Here $M_{tij}$ denotes the corresponding semantic mask.

\noindent\textbf{Total loss for optimization.}
The total loss can be defined as:

\begin{equation}
\mathcal{L}_{total} = \mathcal{L}_{sup}+ \lambda(t)(\mathcal{L}_{psm}+\gamma\mathcal{L}_{mgd}),
\end{equation}
where $\mathcal{L}_{sup} = \mathcal{L}_{cls}+\mathcal{L}_{reg}+\mathcal{L}_{seg}$~\cite{he2017mask}, $\gamma$ is a balanced weight which we set to 5. 
$\lambda(t)$ is a piecewise weight function that guarantees the loss dominated by $\mathcal{L}_{sup}$ at beginning, gradually increases during training, and declines slowly at last.  

\subsection{Implementation Details.}
We use IR-Net~\cite{zhou2019irnet} as our base model, which utilizes instance relations on Mask R-CNN~\cite{he2017mask}.
The Augmentor (A) consists of both color and location transformations.
Specifically, each sample is first randomly adjusted brightness, contrast and Hue, and then conducted random erasing~\cite{zhong2017random}. 
After that, half of them are flipped.
For the first 1000 iterations, only $\mathcal{D}_{L}$ is used. 
The teacher model is initiated by copying the parameters in the student at the 990th iter, which prevents the framework from degenerating by a poor teacher.
We set $\alpha = \min(1-/(t-990),0.99)$ to let the teacher have a larger update rate at the beginning when the student improves quickly. 
During training, each mini-batch includes both labeled and unlabeled images with a ratio of $1:1$.
The  sigmoid-shaped function $\lambda(t)=e^{-125(1-t/1250)^{2}}$ is used for  $t\in[1000,1250]$ and  $\lambda(t)=e^{-12(1- (T-t)/250)^{2}}$ for  $t\in[T-250,T]$, where $T$ is the total iterations.
Pytorch is adopted to implement our framework. The learning rate is initiated to 1e-2 and decayed to 1e-3 and 1e-4 after 5000 and 7000 iterations.  
We adopt SGD algorithm to optimize the network ,and one Titan XP GPU is used for training.
The pseudo code of the proposed MMT-PSM can be found in the supplementary material.

\section{Experiments and Results}

\textbf{Dataset and evaluation metrics.}
The liquid-based Pap test specimen was collected from 82 patients and imaged in $\times40$ resolution with $\sim 0.2529$ \textit{$\mu$m} per pixel. 
This is used as labeled dataset $\mathcal{D}_{L}$ with totally 4439 cytoplasm and 4789 nuclei annotations. 
Then the dataset is divided in patient-level with the ratio of $7:1:2$ for train, valid and test set.
An overlapping ratio of 0.75 is used to crop images into $1000 \times1000$ for the training set, while the valid, as well as the test set, are non-overlapping cropped. In sum, the number of images for train, valid and test is 961, 50 and 98, respectively.
Apart from that, 4371 images from other patients with a resolution of $1000 \times 1000$ are randomly cropped from whole slide images as the unlabeled dataset $\mathcal{D}_{U}$.

We use Average Jaccard Index (AJI)~\cite{kumar2017dataset} and mean Average Precision (mAP)~\cite{he2017mask} for quantitative evaluation.
Results are calculated on cytoplasm (Cyto.), nuclei (Nuc.) and the average (Avg.).
AJI is commonly used in cell nuclei segmentation task, which measures the ratio of the aggregated intersection and aggregated union for all the predictions and ground truths in the image.
mAP is the mean of the average precision under different IOU thresholds, which is widely used in the general detection and instance segmentation tasks.

\begin{table}[t]
\renewcommand\arraystretch{1.2}
    \centering
        \caption{Evaluation of MMT-PSM with different numbers of labeled data.}
    \setlength{\belowcaptionskip}{-10pt}
    \begin{tabular}{p{1.6cm}|p{2.2cm}|p{1.2cm}p{1.2cm}p{1.2cm}|p{1.2cm}p{1.2cm}p{1.2cm}}
    \toprule
      \hfil\multirow{2}{*}{ $\%$ labeled } & \hfil\multirow{2}{*}{Method} & \multicolumn{3}{c|}{AJI$\left [ \% \right ]$} & \multicolumn{3}{c}{mAP$\left [ \% \right ]$} \\
    \cline{3-8}
        &  & \hfil Cyto. &\hfil Nuc. & \hfil Avg. & \hfil Cyto. &\hfil Nuc. & \hfil Avg. \\
        \hline
       
\hfil\multirow{2}{*}{ $10\%$ } &\hfil  IR-Net  &\hfil66.81  &\hfil54.07  & \hfil60.44& \hfil37.36 &\hfil30.14& \hfil 33.75\\

&\hfil MMT-PSM  &\hfil \textbf{ 70.00}&\hfil  \textbf{56.90}& \hfil\textbf{63.45}& \hfil\textbf{39.12} &\hfil\textbf{32.09}& \hfil \textbf{35.61}\\
\hline

\hfil\multirow{2}{*}{ $20\%$ }&\hfil  IR-Net &\hfil70.15  &\hfil54.42  & \hfil62.29& \hfil 40.28&\hfil\textbf{31.75}& \hfil36.02 \\
&\hfil MMT-PSM &\hfil \textbf{71.56} &\hfil \textbf{56.49} & \hfil\textbf{64.03}& \hfil\textbf{41.88} &\hfil31.34& \hfil\textbf{36.61} \\
\hline

\hfil\multirow{2}{*}{ $40\%$ }&\hfil  IR-Net&\hfil70.70 &\hfil\textbf{57.35}  & \hfil64.03& \hfil40.00 &\hfil31.65& \hfil35.83 \\
&\hfil MMT-PSM &\hfil \textbf{71.75} &\hfil57.23  & \hfil\textbf{64.49}& \hfil\textbf{42.30} &\hfil\textbf{32.22}& \hfil\textbf{37.26} \\
\hline

\hfil\multirow{2}{*}{ $80\%$ }& \hfil  IR-Net  &\hfil 72.44 &\hfil 57.72 & \hfil65.08& \hfil 41.35&\hfil33.07& \hfil37.21 \\
&\hfil  MMT-PSM&\hfil \textbf{74.26} &\hfil\textbf{ 59.58} & \hfil \textbf{66.92}& \hfil\textbf{45.49} &\hfil\textbf{34.76}& \hfil \textbf{40.13}\\
\hline

\hfil\multirow{2}{*}{ $100\%$ }&\hfil  IR-Net &\hfil73.38  &\hfil58.38  & \hfil65.88& \hfil43.44 &\hfil32.64& \hfil38.04 \\
&\hfil  MMT-PSM &\hfil\textbf{ 73.45}&\hfil\textbf{60.43} & \hfil\textbf{66.94}& \hfil\textbf{46.01} &\hfil\textbf{35.02}& \hfil\textbf{40.52} \\
\hline

         \toprule
    \end{tabular}
    \label{tab:tb1}
\end{table}

\noindent\textbf{Evaluation on different dataset settings.}
Firstly, we evaluate the impact of leveraging unlabeled images by our proposed model under different amounts of labeled samples. 
Our proposed method (MMT-PSM) is compared with the state-of-the-art fully supervised method, named IR-Net~\cite{zhou2019irnet}, which utilizes the instance relation for mask refinement and duplication removal based on the Mask R-CNN~\cite{he2017mask} structure. 
We evaluate the performance of the proposed MMT-PSM with a varying number of labeled data from 96 to 961 and 4371 unlabeled data. 
The IR-Net is trained with the same labeled data only.
As shown in Table~\ref{tab:tb1}, results from the proposed MMT-PSM achieve relatively consistent improvements on both metrics.
It improves average AJI by $2.98\%$, $1.74\%$, $0.46\%$, $1.84\%$, and $1.06\%$, and also improves average mAP by $1.86\%$, $0.59\%$, $1.44\%$, $2.92\%$, and $2.48\%$ for mAP compared with those only trained on the same number of labeled data, which demonstrates the effectiveness of the proposed SSL method.

\begin{table}[t]
\renewcommand\arraystretch{1.2}
    \centering
        \caption{Quantitative comparisons with state-of-the-arts on the test set.}
    \setlength{\belowcaptionskip}{-10pt}
    \begin{tabular}{p{3cm}|p{1.4cm}p{1.4cm}p{1.4cm}|p{1.4cm}p{1.4cm}p{1.4cm}}
    \toprule
      \hfil\multirow{2}{*}{ Method}& \multicolumn{3}{c|}{AJI$\left [ \% \right ]$} & \multicolumn{3}{c}{mAP$\left [ \% \right ]$} \\
    \cline{2-7}

        &  \hfil Cyto. &\hfil Nuc. & \hfil Avg. & \hfil Cyto. &\hfil Nuc. & \hfil Avg. \\
        \hline

\hfil  IR-Net~\cite{zhou2019irnet}&\hfil73.38  &\hfil58.38  & \hfil65.88& \hfil43.44 &\hfil32.64& \hfil38.04 \\
\hfil  ODKD~\cite{chen2017learning}&\hfil 72.64 &\hfil56.34  & \hfil64.49& \hfil44.23 &\hfil\textbf{35.50}& \hfil39.87 \\
\hfil  FFI~\cite{wang2019distilling}&\hfil \textbf{73.89} &\hfil 59.30 & \hfil66.60& \hfil 44.60&\hfil34.72& \hfil39.66 \\
\hfil\textbf{MMT-PSM}&\hfil 73.45&\hfil\textbf{60.43} & \hfil\textbf{66.94}& \hfil\textbf{46.01 }&\hfil35.02& \hfil\textbf{40.52}\\

         \toprule
    \end{tabular}

    \label{tab:stateoftheart}
\end{table}

\begin{table}[t]
\renewcommand\arraystretch{1.2}
    \centering
        \caption{Quantitative analysis of different components on the test set.}
    \setlength{\belowcaptionskip}{-10pt}
    \begin{tabular}{p{3.6cm}|p{1.3cm}p{1.3cm}p{1.3cm}|p{1.3cm}p{1.3cm}p{1.3cm}}
    \toprule
      \hfil\multirow{2}{*}{ Method}& \multicolumn{3}{c|}{AJI$\left [ \% \right ]$} & \multicolumn{3}{c}{mAP$\left [ \% \right ]$} \\
    \cline{2-7}

        &  \hfil Cyto. &\hfil Nuc. & \hfil Avg. & \hfil Cyto. &\hfil Nuc. & \hfil Avg. \\
        \hline

\hfil  IR-Net~\cite{zhou2019irnet}&\hfil73.38  &\hfil58.38  & \hfil65.88& \hfil43.44 &\hfil32.64& \hfil38.04 \\
\hfil MMT-PSM (w/o $\mathcal{L}_{mgd}$)&\hfil \textbf{75.01} &\hfil59.23 & \hfil\textbf{67.12}& \hfil45.58 &\hfil33.16& \hfil39.37 \\
\hfil  MMT-PSM  (w/o $\mathcal{L}_{psm}$)&\hfil 74.38 &\hfil59.46  & \hfil66.92& \hfil 44.39&\hfil33.75& \hfil39.07 \\
\hfil \textbf{MMT-PSM}&\hfil 73.45&\hfil\textbf{60.43} & \hfil66.94& \hfil\textbf{46.01} &\hfil\textbf{35.02}& \hfil \textbf{40.52} \\

         \toprule
    \end{tabular}

    \label{tab:ablation}
\end{table}

\noindent\textbf{Comparison with other semi-supervised methods.}
We implement and adapt several state-of-the-arts methods for comparison:
1). Chen \etal~\cite{chen2017learning} improved object detection by knowledge distillation (ODKD) with weighted cross-entropy loss for the imbalanced data problem. Meanwhile, feature imitation is conducted in all regions.
2). Wang \etal~\cite{wang2019distilling} proposed fine-grained feature imitation (FFI) which firstly estimated the object anchor locations and then let the student's features be closed to teacher's on the selected regions.
Note that we used the same network backbone~\cite{zhou2019irnet} on these methods with $961$ labeled data and $4371$ unlabeled data for fair comparison.
As can be seen in Table~\ref{tab:stateoftheart}, all the SSL methods outperforms the supervised method on most of the evaluation indicators.
Compared with fully supervised methods, results from ODKD improves $1.85\%$ mAP but decreases AJI. 
The reason is it penalizes the classification and feature discrepancy in all regions.
Therefore it is inevitable to introduce the noise.
FFI selects the proposals closed to objects for feature distillation, hence achieves better results.
Furthermore, the proposed MMT-PSM achieves the best performance over the state-of-the-art SSL methods, illustrating that our method has the keen ability to distillate the information both in feature space and semantic predictions. 

\noindent\textbf{Ablation study of the proposed method.}
We also conduct the ablation study for the impact of proposed components:
1). MMT-PSM (w/o  $\mathcal{L}_{mgd}$) denotes the proposed method without the mask-guided feature distillation, and
2). MMT-PSM (w/o $\mathcal{L}_{psm}$) denotes the proposed method without the perturbation-sensitive sample mining for knowledge distillation.
Results are shown in Table~\ref{tab:ablation}. 
Utilizing perturbation-sensitive samples measured in the teacher network as the pseudo-labels for optimizing the student improves $1.24\%$ for average AJI and $1.33\%$ for mAP.
Meanwhile, forcing features from the teacher and the student being consistent in the foreground region also increases $1.04\%$ average AJI and $1.03\%$ mAP.
Lastly, combining two components in our mean teacher framework achieves the competitive performance by $66.94\%$ AJI and $40.52\%$ mAP.

\begin{figure}[t]
\centering
	\includegraphics[width=0.9\textwidth]{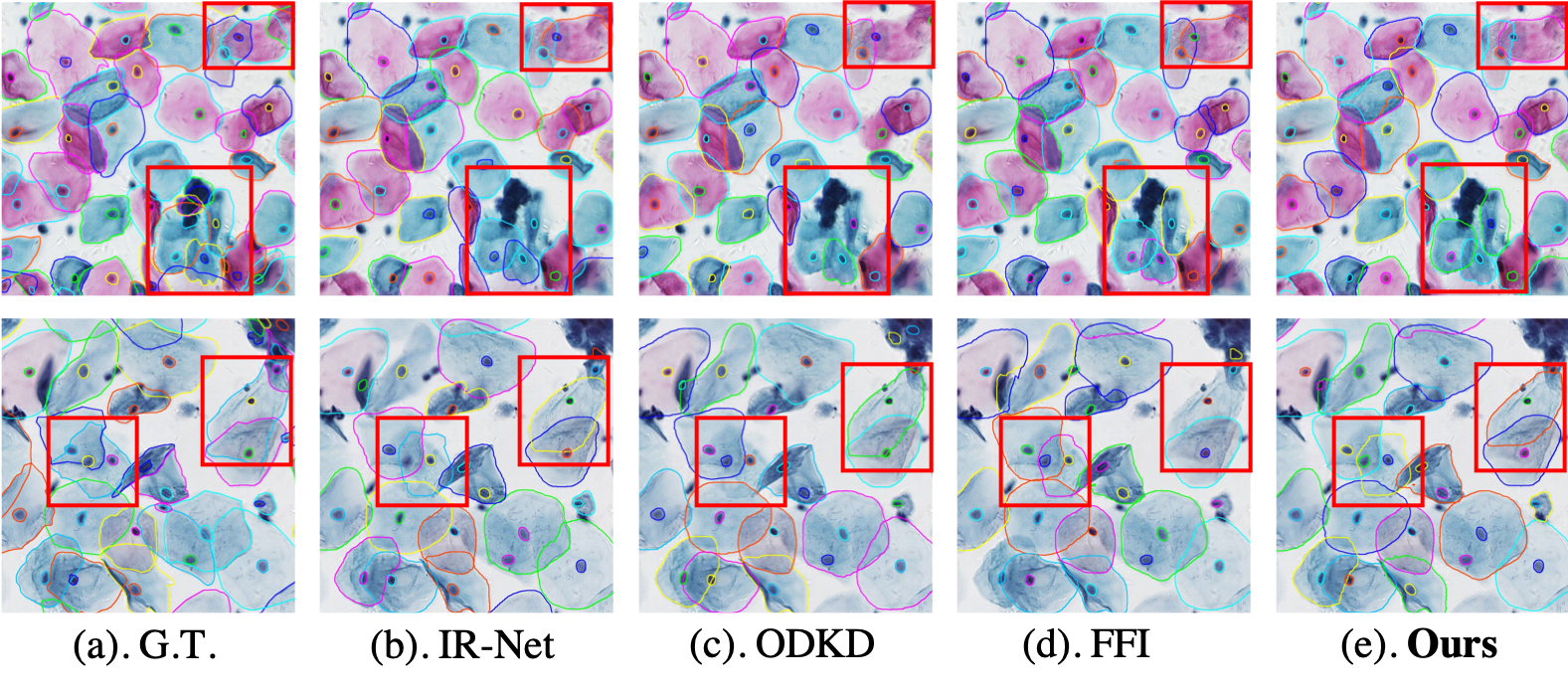}
	\caption{ Qualitative comparisons of semi-supervised cervical cell instance segmentation on the test set. Red rectangles highlight the main difference among different methods.}
		\label{visual}
	
\end{figure}

\noindent\textbf{Qualitative evaluation.}
We also visualize different methods' results from challenging cases including the heavily occlusion of cytoplasm and blurred regions. 
As can be seen in Fig.~\ref{visual}, each closed curve denotes an individual instance.
Compared with other methods, our proposed MMT-PSM has the better ability to recognize the translucent cervical cells in low contrast areas. 

\section{Conclusion}
In this paper, we propose a novel mask-guided mean teacher framework  with  perturbation-sensitive sample mining which conducts knowledge distillation for semi-supervised cervical cell instance segmentation.
The proposed method encourages the network to output consistent feature maps and predictions under small perturbations.
Only samples with high grade of perturbation sensitivity are selected for semantic distillation, which prevents the meaningless guidance from easy background cases.
In addition, the segmentation mask is used as guidance for better feature distillation. 
Experiments demonstrate our proposed method effectively leverage the unlabeled data and outperforms other SSL methods. 
Our proposed MMT-PSM framework is general and can be easily adapted to other semi-supervised medical image instance segmentation tasks.

\section*{Acknowledgments}
The work described in the paper was supported in parts by the following grants from
Key-Area Research and Development Program of Guangdong Province, China (2020B010165004),
Hong Kong Innovation and Technology Fund (Project No. ITS/041/16), National Natural Science Foundation of China (Project No. U1813204) and Shenzhen Science and Technology Program (JCYJ20170413162\\256793).

%
%
    \bibliographystyle{splncs04}
    \bibliography{cite}

\end{document}